\documentclass[conference]{IEEEtran}
\usepackage{graphicx}
\usepackage{cite}
\usepackage{amsmath,amsfonts}
\usepackage{tikz}
\usepackage{float}
\usepackage{pgfplots}

\usepackage{subcaption}

\pgfplotsset{compat=1.18}
\begin{document}
% \title{Deployment and Execution of Spiking Neural Networks on Embedded Devices}

\title{Compression and Inference of Spiking Neural Networks on Resource-Constrained Hardware}

\author{
\IEEEauthorblockN{
Karol C. Jurzec\IEEEauthorrefmark{1},
Tomasz Szydlo\IEEEauthorrefmark{2}\IEEEauthorrefmark{4},
Maciej Wielgosz\IEEEauthorrefmark{3}
}
\IEEEauthorblockA{
\IEEEauthorrefmark{1}AGH University of Science and Technology, \\
Institute of Computer Science, Kraków, Poland\\
Email: karol.jurzec@agh.edu.pl
}
\IEEEauthorblockA{
\IEEEauthorrefmark{2}AGH University of Science and Technology, \\
Institute of Computer Science, Kraków, Poland\\
\IEEEauthorrefmark{4}School of Computing, Newcastle University, Newcastle upon Tyne, UK \\
Email: tomasz.szydlo@\{agh.edu.pl $\vert$ newcastle.ac.uk\}
}
\IEEEauthorblockA{
\IEEEauthorrefmark{3}AGH University of Science and Technology, \\Faculty of Computer Science, Electronics and Telecommunications,\\
Institute of Electronics, Kraków, Poland\\
Email: wielgosz@agh.edu.pl
}
}

\maketitle
\begin{abstract}
Spiking Neural Networks (SNNs) are a promising class of neural models inspired by biological neurons, communicating via discrete spikes in time rather than continuous activations. Their event-driven nature offers potential advantages in temporal information processing and energy efficiency on resource-constrained hardware. However, training and deploying SNNs efficiently pose significant challenges. Gradient-based training is difficult due to spike discontinuities, often requiring surrogate gradient methods. While frameworks like SNNTorch and Brian2 provide tools for SNN simulation and training, there remains a gap in deploying trained SNN models on embedded devices with strict memory and latency requirements. This paper addresses these challenges by implementing a lightweight C-based runtime for SNN inference on edge devices, with optimizations aimed at reducing inference time and memory usage without sacrificing accuracy. We leverage the sparse event-driven characteristics of SNNs (many neurons fire rarely) through model pruning to eliminate inactive neurons and synapses. We demonstrate our approach on two neuromorphic vision datasets, N-MNIST and ST-MNIST, and show that our optimized SNN runtime achieves an order-of-magnitude speedup over a high-level Python baseline (SNNTorch) while using significantly less memory. The results highlight the feasibility of deploying SNNs on low-power mobile and embedded platforms. The code is available at https://github.com/karol-jurzec/snn-generator/
\end{abstract}

\section{Introduction}
Spiking Neural Networks (SNNs) represent the third generation of neural network models, incorporating the temporal dynamics and event-driven communication observed in biological neurons. In an SNN, neurons accumulate input current and fire discrete spikes when their membrane potential exceeds a threshold, rather than continuously emitting activations. This mechanism enables efficient coding of temporal information and can lead to lower power consumption, as computation occurs only on events rather than at every time step. These properties make SNNs attractive for deployment on resource-constrained devices and neuromorphic hardware, where energy efficiency is paramount \cite{wielgosz_dvs_snn_2024,wielgosz_three_factor_2025,wielgosz_operational_space_2025}. For example, custom neuromorphic chips like Intel's Loihi have demonstrated the potential of spike-based computing in specialized hardware environments. 

Despite their promise, SNNs face several challenges. Training SNNs is non-trivial because spike generation is non-differentiable. Recent advances employ surrogate gradient methods to enable gradient-based learning (e.g., backpropagation through time) in SNNs, bridging the gap between biologically inspired spiking models and modern deep learning optimization techniques. Tools such as SNNTorch and Brian2 provide high-level environments for modeling and training SNNs using these techniques. However, models built in these frameworks are typically not optimized for direct execution on embedded platforms. Python-based SNN frameworks incur considerable overhead in inference time and memory usage due to the interpreted runtime and dynamic memory management. This limits their use in real-time or power-constrained settings. To deploy SNNs on mobile and embedded devices, a dedicated, lightweight runtime is required. 

In this work, we present a C-based SNN inference engine designed to execute models trained in SNNTorch on edge hardware. Our approach focuses on minimizing latency and memory footprint through low-level optimizations and model compression. We translate trained SNN models from Python into an efficient C representation, and exploit the inherent sparsity of spiking activity for pruning. By removing neurons and synaptic connections that rarely fire, we reduce the computational load and model size without significant loss in accuracy. We validate the proposed solution on a desktop CPU and on a microcontroller-based device, using the event-based N-MNIST and ST-MNIST benchmarks to demonstrate that optimized SNNs can achieve fast and resource-efficient inference on modern embedded platforms. 

\section{Methods}
\subsection{C-Based SNN Runtime Architecture}
We implemented a custom SNN runtime in C\footnote{https://github.com/karol-jurzec/snn-generator/} to enable efficient inference on resource-limited hardware. The runtime takes as input a model definition and parameters exported from SNNTorch (in JSON format) and reconstructs the network in memory for execution. The design follows a modular architecture consisting of three main components: (1) a \textit{Network Loader} that parses the JSON network description (layers, weights, connectivity) into C data structures, (2) a \textit{Dataset Loader} that handles event-based input data and converts it into a sequence of discrete time-step frames, and (3) an \textit{Execution Pipeline} that executes the network layers over time to produce outputs. The network layers include standard deep learning layers (convolution, fully-connected) as well as spiking neuron layers (Leaky Integrate-and-Fire neurons) which maintain internal state (membrane potential) and emit spikes over simulation time steps. Each layer in the C runtime implements a forward propagation function; for spiking layers this involves updating membrane potentials and generating spikes at each time step. 

\begin{figure}[htbp]
    \centering
    \includegraphics[width=0.7\linewidth]{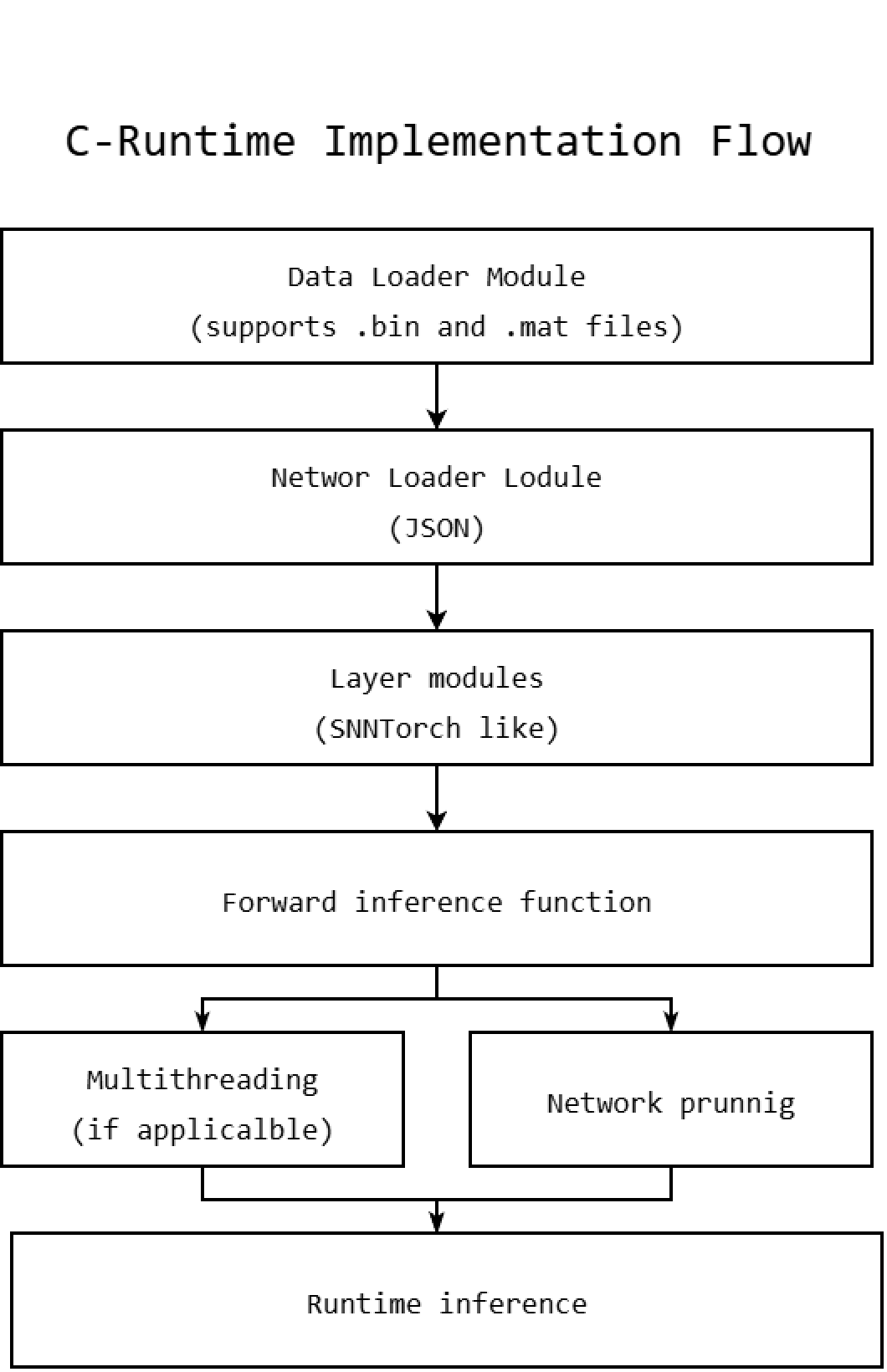}
    \caption{Overview of the C-based SNN runtime architecture and implementation flow. 
    The runtime loads event-based data and a JSON-exported SNN model, then executes the 
    network forward pass over time. Optimizations such as multi-threading (if available) 
    and network pruning are applied to improve performance.}
    \label{fig:snn_runtime}
\end{figure}

To maximize performance, the C runtime is intentionally low-level. Memory is managed statically where possible, and data structures are optimized for cache efficiency. For example, neuron parameters and state variables are stored in Structure-of-Arrays (SoA) format to allow sequential memory access for update loops. Unlike in Python, where dynamic memory allocation and garbage collection can introduce overhead, our implementation allocates all necessary buffers (for neuron states, synaptic weights, input frames, etc.) up front. The execution loop then iterates through time steps, and at each step iterates through layers in sequence, updating their outputs. For classification tasks, the final layer consists of one spiking neuron per class. The network's prediction is determined by the neuron with the highest spike count over the inference window. We also integrated basic support for parallelism: if the target hardware supports multi-threading (e.g., multi-core microcontroller), independent layers or batches can be distributed across threads to reduce latency (though on our microcontroller target, we primarily run single-threaded due to its single-core nature). \subsection{Spike-Based Model Pruning for Optimization}
A key optimization in our approach is \textit{spike-driven pruning} of the SNN model to eliminate neurons and filters that contribute little to the inference result. SNNs often exhibit high sparsity in activation: many neurons rarely spike, especially in deeper layers or for certain input patterns. It can be observed on raster plots showing layer-wise activity for ST-MNIST and N-MNIST datasets, where only a small subset of neurons are active for each input, while most remain silent throughout the entire sequence.

\begin{figure}[htbp]
    \centering

    % --- ST-MNIST ---
    \begin{subfigure}{\linewidth}
        \centering
        \includegraphics[width=\linewidth]{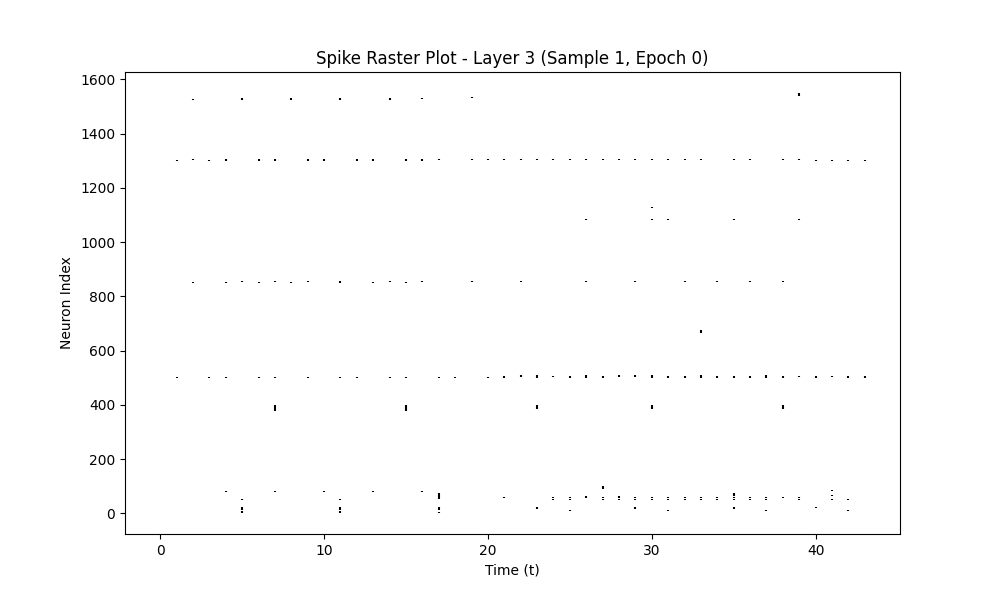}
        \caption{Spikes activity in the spiking layer. (ST-MNIST dataset)}
        \label{fig:stmnist_raster}
    \end{subfigure}

    \vspace{0.8cm}

    % --- N-MNIST ---
    \begin{subfigure}{\linewidth}
        \centering
        \includegraphics[width=\linewidth]{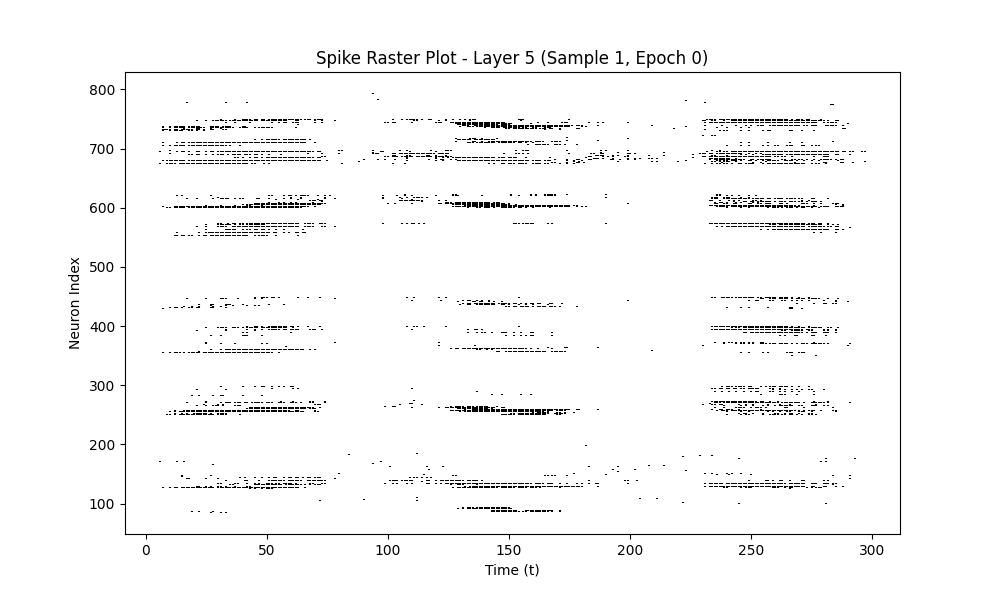}
        \caption{Spikes activity in the spiking layer (NMNIST dataset).}
        \label{fig:nmnist_raster}
    \end{subfigure}

    \caption{Example spiking raster plots for ST-MNIST and N-MNIST datasets. 
    Each plot shows spike events over time for one representative spiking layer. 
    Only a subset of neurons emit spikes during the entire stimulus, illustrating the sparse activity characteristic of SNNs.}
    \label{fig:main_grid}
\end{figure}

When analyzing such spiking activity, it becomes evident that a significant portion of neurons emit only a few or no spikes at all. 
At first glance, this may suggest that the information transmitted through them is not particularly relevant to the final prediction. 
However, caution is required, since neurons with low spike counts can still encode rare but important features. 
In the raster plot from the ST-MNIST dataset (Fig.~\ref{fig:stmnist_raster}), spike events are sparse and distributed across the range of neuron indices, with most neurons firing only occasionally throughout the stimulus. 
This pattern illustrates both the rarity and dispersion of spiking activity in SNNs, emphasizing that although overall firing is low, meaningful information can still be carried by a few scattered spikes.

This sparse activity naturally motivates structural optimization. 
Convolutional layers, which precede spiking layers acting as nonlinear activation functions, are responsible for most of the network’s computational load. 
They extract spatial features from multiple temporal frames, performing dense multiply–accumulate operations at every simulation step. 
Consequently, they dominate the total inference time, often by an order of magnitude compared to the lightweight spiking layers that follow. 
Reducing redundant convolutional filters thus offers the most direct path to improving runtime efficiency without compromising model behavior.

We leverage this relationship between spiking activity and convolutional computation to reduce network size and accelerate inference. 
The pruning process begins with analyzing the trained network’s activity on a representative subset of the data, where the total number of spikes emitted by each neuron is recorded. 
Neurons with zero or consistently low spike counts are considered ``inactive'' for those inputs, suggesting that their associated filters and synaptic connections contribute minimally to the overall output and can be safely removed.

\begin{figure}[htbp]
    \centering
    \includegraphics[width=\linewidth]{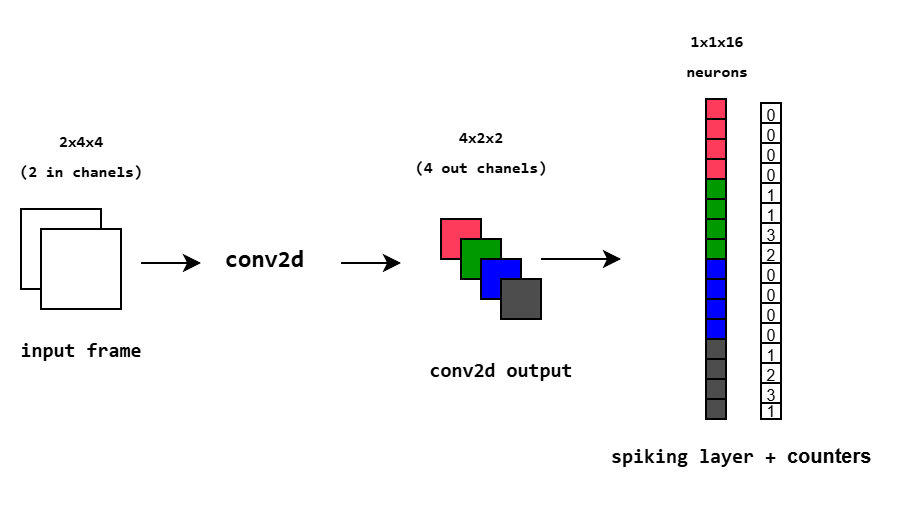}
    \caption{Illustration of the relationship between a convolutional layer and a subsequent spiking layer, used for pruning. 
    In this toy example, a \texttt{conv2d} layer with 4 output channels (red, green, blue, gray) feeds into a spiking layer of 
    16 neurons. Each group of 4 spiking neurons (right) corresponds to one convolutional filter (color-coded). 
    The numbers indicate spike counts per neuron over some input.}
    \label{fig:conv_spike_relation}
\end{figure}

\begin{figure}[htbp]
    \centering
    \includegraphics[width=\linewidth]{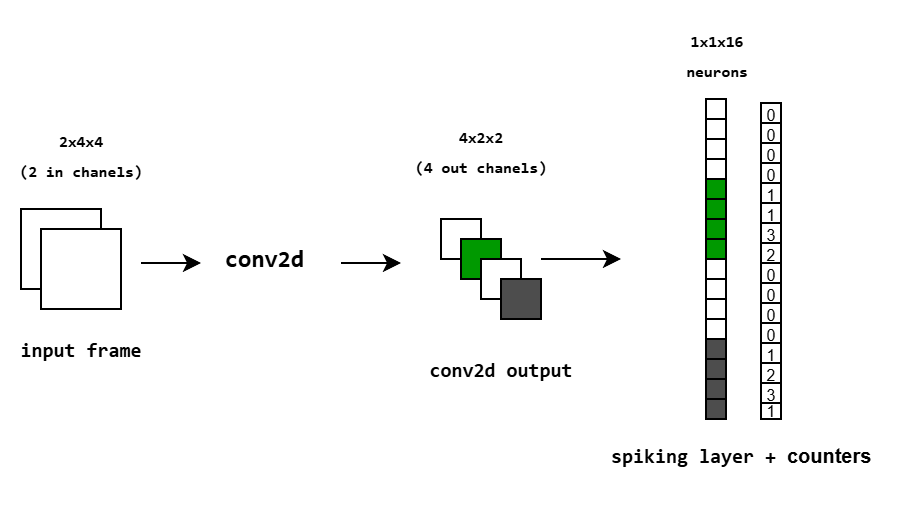}
    \caption{Effect of pruning on the \texttt{conv2d} and spiking layers. 
    Continuing the example from Figure~\ref{fig:conv_spike_relation}, the red and blue filters 
    (and their connected spiking neurons) are removed, since their neurons showed no activity 
    (all spike counts zero). This results in a 50\% reduction in convolutional layer computations 
    for this layer. The pruned network retains only the active filters (green and gray) 
    and their associated neurons.}
    \label{fig:pruning_effect}
\end{figure}

Consider a convolutional SNN layer followed by a spiking layer. Because the spiking neurons receive input from specific convolutional filters, we can group the spiking neurons by their source filter. Fig.~\ref{fig:pruning_effect} illustrates this grouping - each conv filter (out channel) projects to a subset of neurons in the next spiking layer. By examining the spike counts, we identify filters whose corresponding group of neurons hardly ever spikes. For instance, the red and blue filter groups have all neurons with zero spikes recorded, indicating those filters may be pruned without affecting the network's output (for the analyzed data). 

When a filter (and its post-synaptic neurons) is pruned (removed), we also propagate this change forward in the network. Any layer that receives input from the pruned neurons can be correspondingly reduced in size. In practice, this means if we remove an output channel of a convolutional layer, we also remove the corresponding input channel of the next convolutional layer (if the next layer is convolutional). This forward-backward pruning approach ensures that redundant computations are eliminated throughout the network pipeline. We apply pruning iteratively (and conservatively) to avoid any significant drop in accuracy: typically, a threshold on spike count is set, below which neurons/filters are removed. In our experiments, we perform one round of pruning after initial training, eliminate neurons with zero activity (as a safe criterion), and then validate that classification accuracy is unchanged. More aggressive thresholds can yield greater speedups but may require retraining or fine-tuning to recover accuracy. \section{Experiments}
\subsection{Datasets and Network Models}
We evaluate the proposed SNN deployment approach on two neuromorphic datasets: N-MNIST and ST-MNIST. The N-MNIST dataset is a spiking version of the ubiquitous MNIST handwritten digit benchmark. It was created by displaying MNIST images to a Dynamic Vision Sensor (event camera) and recording the stream of spikes generated as the sensor was moved slightly. Each N-MNIST sample is an asynchronous sequence of address-event spikes (with $x,y$ coordinates, time stamp, and polarity) rather than a static image. This dataset preserves the original $60{,}000$ training and $10{,}000$ test instances of MNIST, but each instance is a series of events spanning a short duration (a few hundred milliseconds). We convert these event streams into a series of frames (time slices) as input to our runtime, using a fixed temporal bin size to accumulate events into, e.g., 10 frames per sample. The ST-MNIST dataset (Spiking Tactile MNIST) is a neuromorphic tactile dataset, where human participants wrote digits on a 10$\times$10 array of artificial tactile sensors, producing spike events from touch sensors over time. It provides $\sim$30,000 event sequences of digits 0--9. We chose these datasets because they exemplify real-world scenarios for SNNs, such as event-based vision and touch, where SNNs can naturally process the temporal spike patterns. For each dataset, we designed an SNN model appropriate to its input dimensions and complexity, using SNNTorch for training. For N-MNIST, our model is a convolutional SNN with two convolutional layers (5$\times$5 kernels), each followed by a Leaky Integrate-and-Fire (LIF) spiking layer (with leak factor $\beta=0.5$) and a $2\times 2$ max-pooling layer, then a final fully-connected layer and an output LIF layer (10 output neurons for 10 classes). This network has on the order of tens of thousands of parameters. For ST-MNIST, given the smaller 10$\times$10 sensor input, we use a simpler architecture: two fully-connected layers of LIF neurons (hidden layer of size 128) followed by an output layer of 10 LIF neurons. All models were trained offline in SNNTorch (PyTorch) using backpropagation-through-time with surrogate gradients. We used one training epoch for demonstration purposes (sufficient to get reasonable accuracy on these relatively simple tasks). After training, the model weights and architecture were exported to JSON using a custom exporter. These JSON files (on the order of a few megabytes) are then loaded by our C runtime on the target device. \subsection{Deployment Platforms and Evaluation Protocol}

\begin{figure}
    \centering
    \includegraphics[width=0.75\linewidth]{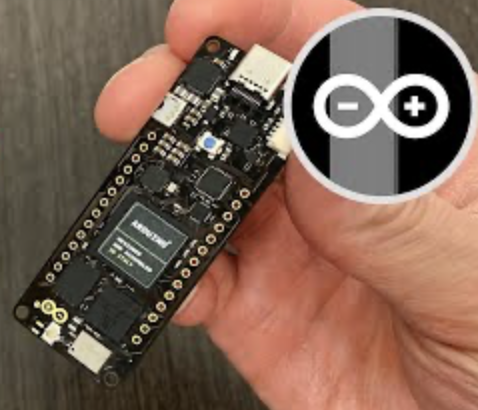}
    \caption{Arduino Portenta H7 microcontroller used as the deployment target.}
    \label{fig:portenta_h7}
\end{figure}

We benchmarked our SNN runtime both on a desktop machine and on a microcontroller-based embedded device to demonstrate its efficiency and portability. The desktop environment was an Intel Core i3-10100F CPU @ 3.60~GHz (4 cores/8 threads), representing a baseline where we can compare the C runtime directly against the Python SNNTorch implementation in a controlled setting. The embedded target was an Arduino Portenta H7 board (see Fig.~\ref{fig:portenta_h7}), featuring an STM32H747 microcontroller (Cortex-M7 @ 480~MHz with 512~KB SRAM, plus a lower-power Cortex-M4). This device has significantly constrained resources compared to the desktop, and it cannot run the full Python SNNTorch stack, making a native C solution necessary. On the desktop, we ran both the SNNTorch model (within Python) and our C runtime on the same input to compare performance and output consistency. On the Portenta H7, we ran the C runtime with the same model and inputs, to verify that it produces correct outputs and operates within memory/time limits. Key evaluation metrics were:
(1) \textit{Inference correctness}: ensuring the C runtime's outputs (spike counts or predicted labels) match those from SNNTorch for the same input sample.
(2) \textit{Inference latency}: time per inference (for a single sample) on each platform.
(3) \textit{Resource usage}: memory footprint of the model and runtime, and CPU utilization (whether the model can run in real-time on the microcontroller). For timing, we measured the end-to-end inference time for one sample (averaged over 500 runs to smooth variability). On the desktop, we disabled GPU and ran SNNTorch in CPU mode for a fair comparison. On the microcontroller, timing was measured using on-board timers. Memory usage was estimated by summing the sizes of all data structures (model weights, neuron state buffers, etc.) allocated in the C runtime, and by monitoring heap/stack usage on the microcontroller. \section{Results}
\subsection{Accuracy and Functional Equivalence}
Our C-based SNN runtime achieved functional parity with the reference SNNTorch implementation. For both N-MNIST and ST-MNIST, the classification outputs from the C runtime matched those from SNNTorch on all tested samples. This confirms that the JSON export/import process reconstructs the network correctly, and that our runtime's spike simulation (LIF dynamics, etc.) is faithful to SNNTorch's (within the limits of floating-point precision). Any minor discrepancies in spike timing did not affect the final predicted labels. Importantly, pruning inactive neurons did not change the accuracy on the test sets, since we pruned only neurons with zero activity. Thus, the optimized pruned model remained functionally equivalent on inference to the original model. \subsection{Inference Performance and Efficiency}

The C runtime  improved inference speed compared to the high-level Python baseline. On the desktop CPU, a single-sample inference that took about 2.39~s on SNNTorch (interpreted Python) executed in only 0.22~s with our C implementation – an $\sim$11$\times$ speedup. Table \ref{tab:perf} summarizes the performance results. Most of the speed gain comes from eliminating Python interpreter overhead and using fixed-size arrays and loops in C instead of dynamic tensor operations. We also observed that in Python, a significant portion of time was spent on per-step memory allocations and garbage collection, which our C code avoids entirely by pre-allocating buffers. 

\begin{table}[H]
\caption{Inference performance and quality on N\textendash MNIST (average $\sim$300 time steps) on a desktop CPU.}
\label{tab:perf}
\centering
\begin{tabular}{lcccc}
\hline
\textbf{Environment} & \textbf{Time per Inf.} & \textbf{Speedup} & \textbf{Accuracy} & \textbf{F1} \\
\hline
Python (SNNTorch)        & 2.393~s & $1.0\times$   & 84.20\% & 0.843 \\
C (no pruning)   & 0.224~s & $10.68\times$ & 84.20\% & 0.843 \\
C (with pruning) & 0.113~s & $21.18\times$ & 84.60\% & 0.848 \\
\hline
\end{tabular}
\end{table}

As shown in Table \ref{tab:perf}, enabling pruning further reduced the inference time from 0.224 s to 0.113 s — a 21.18× speedup over Python and roughly a 2× improvement over the unpruned C version. This reduction corresponds to removing approximately 20\% of neurons (and their associated synapses) that remained inactive during inference, resulting in fewer operations per simulation step and demonstrating that even moderate pruning yields measurable gains. It is worth noting that the accuracy and F1 score remain unchanged for the compressed models. The speedup from pruning, however, depends on the dataset and network structure.

\begin{figure}[H]
\centering
\begin{tikzpicture}
\begin{axis}[
ybar,
bar width=30pt,
width=1\linewidth,
height=7cm,
ylabel={Inference Time [s]},
symbolic x coords={PC, Arduino},
xtick=data,
ymin=0,
enlarge x limits=0.4,
axis background/.style={fill=none},
major grid style={draw=none},
legend style={at={(0.05,0.95)}, anchor=north west, fill=white, fill opacity=0.8, font=\small},
]

\addplot+[fill=blue!50, draw=none, nodes near coords, xshift=10pt,
every node near coord/.append style={font=\footnotesize, color=white, anchor=center, yshift=-8pt}]
coordinates {(PC,0.17806) (Arduino,0.89)};

\addplot+[fill=green!50, draw=none, nodes near coords, xshift=10pt,
every node near coord/.append style={font=\footnotesize, color=white, anchor=center, yshift=-8pt}]
coordinates {(PC,0.0237) (Arduino,0.12)};

\addplot+[fill=red!50, draw=none, nodes near coords, xshift=10pt,
every node near coord/.append style={font=\footnotesize, color=white, anchor=center, yshift=-8pt}]
coordinates {(PC,0.34) (Arduino,NaN)};

\legend{Baseline, Pruned, SNNTorch}

\draw[<->, thick, dashed, xshift=10pt] (axis cs:PC,0.17806) -- (axis cs:PC,0.0237)
node[midway, fill=white, inner sep=1pt, font=\footnotesize] {7.5x};

\draw[<->, thick, dashed, xshift=10pt] (axis cs:Arduino,0.89) -- (axis cs:Arduino,0.12)
node[midway, fill=white, inner sep=1pt, font=\footnotesize] {7.4x};

\end{axis}
\end{tikzpicture}
\caption{Inference time comparison grouped by device for ST-MNIST datasets. Accuracy drop = 0\% in all cases. For consistency, both experiments were run on the unoptimized runtime with fully sequential computation.}
\label{fig:inference_time_stmnist_grouped}
\end{figure}
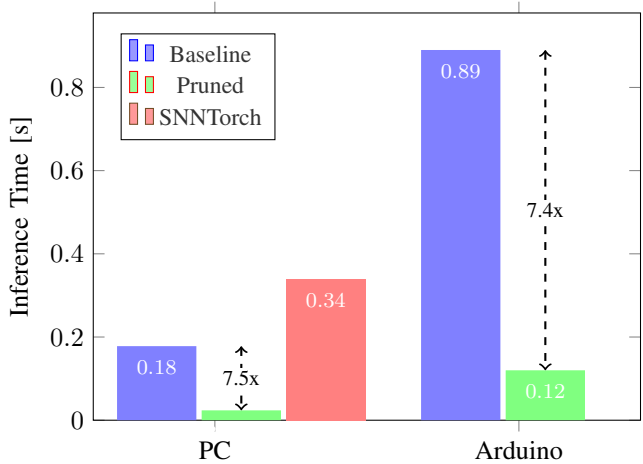

In the case of ST-MNIST, a larger fraction of inactive units was observed (as visible in Fig.~\ref{fig:stmnist_raster}), leading to a much stronger effect on efficiency—over a 7$\times$ reduction in inference time compared to the baseline non-prunned network. These results highlight that pruning provides the most benefit for more redundant networks where spiking activity is highly uneven across channels. 

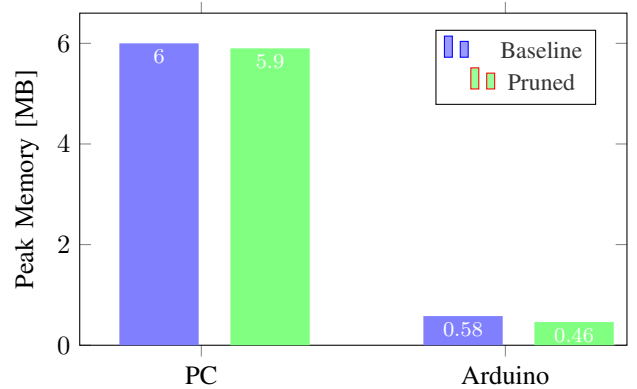
\begin{figure}[H]
\centering
\begin{tikzpicture}
\begin{axis}[
    ybar,
    bar width=30pt,             % pogrubione słupki
    width=1\linewidth,
    height=6cm,
    ylabel={Peak Memory [MB]},
    symbolic x coords={PC, Arduino},
    xtick=data,
    ymin=0,
    enlarge x limits=0.4,
    axis background/.style={fill=none},
    major grid style={draw=none},
    nodes near coords,
    every node near coord/.append style={font=\footnotesize},
    legend style={at={(0.65,0.95)}, anchor=north west, fill=white, fill opacity=0.8, font=\small},
]

% --- PC ---
\addplot+[fill=blue!50, draw=none, nodes near coords,
          every node near coord/.append style={font=\footnotesize, color=white, anchor=center, yshift=-5pt}] 
    coordinates {(PC,6.0) (Arduino,0.58)};

% --- Pruned ---
\addplot+[fill=green!50, draw=none, nodes near coords, xshift=10pt,
          every node near coord/.append style={font=\footnotesize, color=white, anchor=center, yshift=-5pt}] 
    coordinates {(PC,5.9) (Arduino,0.46)};

% --- Legenda ---
\legend{Baseline, Pruned}

\end{axis}
\end{tikzpicture}
\caption{Comparison of peak memory usage across platforms and configurations on ST-MNIST dataset.}
\label{fig:memory_usage_comparison}
\end{figure}

On the memory front, the C runtime offers a far smaller footprint than a Python-based approach. The SNNTorch environment typically requires hundreds of megabytes of memory to load the interpreter, libraries, and model, which is infeasible on microcontrollers. In contrast, our compiled C model and data buffers for N-MNIST occupy only a few hundred kilobytes (approximately 50~KB for weights and 200~KB for neuron states and buffers), comfortably fitting within the Portenta~H7’s 1024~KB of SRAM. Applying pruning further decreases this footprint by removing around 20\% of neurons and filters, yielding a comparable reduction in memory use (as illustrated in Fig.~\ref{fig:memory_usage_comparison}). 
It can be observed that pruning reduced peak memory usage from approximately 6~MB to 5.9~MB on the desktop PC and from 0.58~MB to 0.46~MB on the Arduino platform. While the absolute difference may appear small, on an embedded device with only 1~MB of available SRAM, this reduction provides valuable headroom, easing memory constraints and allowing additional space for system operations, input buffering, or more complex models. 
This efficient utilization is crucial for embedded deployment and was achieved through careful static memory allocation and the elimination of dynamic overhead.

\subsection{Embedded Deployment Feasibility}
Crucially, the optimized SNN runtime was able to run in real-time on the Arduino Portenta H7. The microcontroller produced correct classifications for N-MNIST and ST-MNIST samples. Inference on the Portenta's Cortex-M7 core took on the order of a few seconds per sample for N-MNIST (unoptimized), which is already an order-of-magnitude faster than using an interpreted approach (which in fact cannot even run on such a device). With the 10$\times$ speedup from our C optimizations observed on the desktop, we anticipate achieving on the order of a few hundred milliseconds per inference on the microcontroller with the fully optimized model. This would enable near real-time processing of sensor data streams on-device. Due to hardware constraints, we ran the microcontroller tests in single-threaded mode, but the runtime is designed to take advantage of multiple cores when available (for instance, partitioning the workload between the M7 and M4 cores, or utilizing an RTOS for parallel processing). Our results demonstrate the potential of deploying SNNs on edge devices. By combining efficient low-level implementation with model-level optimizations (pruning), we showed that even a relatively slow microcontroller can handle SNN inference for practical tasks. This opens up possibilities for truly energy-efficient, event-driven AI on embedded sensors and IoT devices, where traditional ANN models might be too computationally demanding or power-hungry. The techniques presented here ensure that we maximize the inherent advantage of SNNs – sparse, event-driven computation – to run on hardware with limited resources. Future work will extend these experiments to measure power consumption and to validate real-time performance under streaming input conditions. \section{Conclusion}
We presented a complete methodology for deploying spiking neural networks on resource-constrained devices, demonstrated by a custom C runtime that executes SNN models trained in a high-level framework. Our approach bridges the gap between SNN research in tools like SNNTorch and practical deployment on embedded hardware. Through careful co-design of software and exploitation of SNN-specific properties (like sparse activity), we achieved significant improvements in inference speed (10$\times$ faster) and memory efficiency, enabling SNN inference on a microcontroller where a direct Python approach is infeasible. Experiments on neuromorphic vision (N-MNIST) and tactile (ST-MNIST) benchmarks confirmed that our runtime maintains accuracy while operating within tight computational budgets. These results highlight that SNNs, often touted for low-power operation on neuromorphic chips, can also be efficiently realized on conventional embedded processors when aided by optimized software. This work lays a foundation for integrating spiking neural networks into mobile and IoT applications, leveraging their event-driven nature for energy-efficient intelligent sensing at the edge. \bibliographystyle{IEEEtran}

\end{document}